\pdfoutput=1
\documentclass[11pt]{article}

\usepackage[preprint]{acl}

\usepackage{times}
\usepackage{latexsym}
\usepackage{multirow}
\usepackage{float}
\usepackage{stfloats}

\usepackage[T1]{fontenc}
\usepackage[utf8]{inputenc}

\usepackage{microtype}
\usepackage{algorithm}
\usepackage{algpseudocode}
\usepackage{inconsolata}
\makeatletter
\newcommand{\forcefloats}{%
  \@ifnextchar[H]%
    {\@float}%
    {\@float[H]}%
}

\usepackage{graphicx}
\usepackage[dvipsnames]{xcolor}

\hypersetup{
    colorlinks=true,
    linktoc=all,    
    linkcolor=black, 
}

\hypersetup{
    colorlinks=true,
    linktoc=all,    
    linkcolor=black, 
}

\usepackage[skip=1pt]{caption}

\title{Not What the Doctor Ordered: Surveying LLM-based De-identification and Quantifying Clinical Information Loss}

\author{
Kiana Aghakasiri\textsuperscript{1} \quad
Noopur Zambare\textsuperscript{1} \quad
JoAnn Thai\textsuperscript{2} \\[0.6em]
\textbf{Carrie Ye\textsuperscript{2,4}} \quad
\textbf{Mayur Mehta\textsuperscript{2}} \quad
\textbf{J. Ross Mitchell\textsuperscript{1,2,3}} \quad
\textbf{Mohamed Abdalla\textsuperscript{1,2,3}}
\\[0.6em]
\textsuperscript{1}Department of Computing Science, University of Alberta \\
\textsuperscript{2}Department of Medicine, University of Alberta\\
\textsuperscript{3}Alberta Machine Intelligence Institute (Amii)\\
\textsuperscript{4}Arthritis Research Canada\\[0.6em]
\texttt{{kaghakas, zambare, jjthai, cye, mayur1, jmitche2, mabdall2}@ualberta.ca}
}

\begin{document}

\maketitle
\begin{abstract}
De-identification in the healthcare setting is an application of NLP where automated algorithms are used to remove personally identifying information of patients (and, sometimes, providers). With the recent rise of generative large language models (LLMs), there has been a corresponding rise in the number of papers that apply LLMs to de-identification. Although these approaches often report near-perfect results, significant challenges concerning reproducibility and utility of the research papers persist. This paper identifies three key limitations in the current literature: inconsistent reporting metrics hindering direct comparisons, the inadequacy of traditional classification metrics in capturing errors which LLMs may be more prone to (i.e., altering clinically relevant information), and lack of manual validation of automated metrics which aim to quantify these errors. To address these issues, we first present a survey of LLM-based de-identification research, highlighting the heterogeneity in reporting standards. Second, we evaluated a diverse set of models to quantify the extent of inappropriate removal of clinical information. Next, we conduct a manual validation of an existing evaluation metric to measure the removal of clinical information, employing clinical experts to assess their efficacy. We highlight poor performance and describe the inherent limitations of such metrics in identifying clinically significant changes. Lastly, we propose a novel methodology for the detection of clinically relevant information removal.
\end{abstract}

\section{Introduction}
Free-text clinical notes hold significant potential for NLP applications in healthcare, as they contain extensive patient information crucial to providing care and conducting research. These notes often include personally identifiable information (PII) such as names, addresses, and other identifiers, leading to privacy concerns that frequently restrict research access. Removing PII can mitigate these concerns, reduce the risk of privacy breaches, and enable the use of this valuable clinical information for research purposes. The process of de-identification, which involves removing PII, is often required by regulations such as the Health Insurance Portability and Accountability Act (HIPAA) and the General Data Protection Regulation (GDPR).
\begin{figure}[t]
\includegraphics[width=\columnwidth]{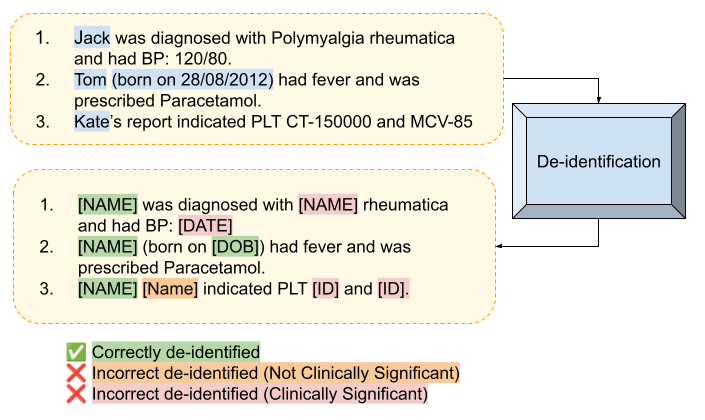}
\caption{Demonstration of correct and incorrect de-identification. Incorrect de-identification my change clinically relevant facts.
\label{fig:demo}}
\end{figure}

Traditional de-identification approaches have evolved from using rule-based and dictionary-based methods to supervised machine learning models trained on large annotated datasets \citep{abdalla2022rethinking}. Although effective (i.e., recall in the mid- to high-90s) in the contexts on which they are trained, these algorithms often do not generalize well to notes from other contexts (e.g., institutions, specialties, etc.) \citep{ferrandez2012generalizability, yang2019study, chen2024examining}, requiring significant manual effort to train for each specific context,  and often struggle in handling non-standard PII (e.g., rare abbreviations) \citep{dernoncourt2017identification}. 

Recent advances in transformer-based models \citep{vaswani2017attention} such as BERT \citep{devlin2019bert}, ClinicalBERT \citep{huang2019clinicalbert}, BioBERT \citep{lee2020biobert} and generative large language models (LLMs) such as Llama \cite{grattafiori2024llama}, ChatGPT \citep{radford2018improving} and Mixtral \cite{jiang2024mixtral} have shown promise in overcoming these limitations. LLMs are pre-trained on very large datasets, giving better generalization without retraining across multiple contexts \citep{radford2019language, kim2024generalizing}. The few-shot and zero-shot capabilities of LLMs can mitigate the need for extensive manual annotation and task-specific training \citep{brown2020language}. Unlike traditional token-based or rule-based systems, LLMs can encode semantic meaning in context, handling rare PII more robustly. As a result, many recent works have proposed the use of LLMs for de-identification and claimed to achieve state-of-the-art performance with relatively minimal effort. 

However, claims of state-of-the-art performance are hard to substantiate. In this work, we conducted an in-depth review of LLMs used to de-identify clinical notes in the NLP literature. We found that models are often evaluated on different datasets, different cuts of the same data, and there is little agreement on which metrics are reported. These limitations hinder direct comparisons, which reduces the impact of these works. Some works do not provide their final prompts further reducing the reproducibility of the work.

In addition to inconsistent performance reporting, only one previous work \cite{pissarra2024unlocking} evaluated `Clinical Information Retention' (CIR): ``the impact of anonymization on the preservation of clinical concepts'' \cite{pissarra2024unlocking}, illustrated in Figure \ref{fig:demo}. Although this issue is not novel (i.e., it also affects traditional de-identification methods), it was brought to researchers' attention given the observed hallucinations of LLMs. Recognizing the importance of this, \citet{pissarra2024unlocking} proposed a novel metric to measure CIR. However, the metric has not yet been manually validated.

Three issues limit our ability to accurately assess the performance of LLM-based de-identification: 1) lack of standardized evaluation data, methodology, and metrics; 2) lack of evaluation of CIR; and 3) use of nonvalidated metrics for CIR.   

The contributions in this paper help address these issues and can be summarized as follows:
\begin{itemize}
    \item Review of the literature on LLM-based de-identification, highlighting inconsistent evaluation practices and their limitations.
    \item Introducing the concept of high and low severity for clinically relevant changes.
    \item Manual validation of metrics previously proposed in the literature (highlighting their severe limitations).
    \item Development of a novel manually validated evaluation metric to capture clinically significant changes introduced by LLMs.
\end{itemize}

\section{Related Work}
\subsection{De-identification (Pre-LLMs)}
Traditional de-identification techniques include dictionary-based (using a predefined list of identifiers to flag PII) \citep{thomas2002successful} and statistical replacement (using classical machine learning or neural networks \citep{liu2017identification}). These techniques work by searching for PII and removing or replacing it. Such approaches often fail to correctly identify all PII and much research has been done to improve them \citep{abdalla2020using}. 
% While traditional NLP-based \citep{liu2017identification} and rule-based methods \citep{menger2018deduce} rely on precision and recall for evaluation, they often miss sensitive information, causing high privacy risks. \citet{abdalla2020using} addressed this by replacing each token in clinical notes with similar words based on their embeddings. 

\subsection{De-identification (LLMs)}
To address the limitations of the previously existing approaches, researchers explored applying transformer-based models like BERT and its variants (ClinicalBERT, BioBERT, RoBERTa \citep{liu2019roberta}), which use contextual embeddings to improve performance over existing neural network approaches, to the task of de-identification. The BERT-based model achieved accuracy, precision and an F1 score in the high 90s \citep{johnson2020deidentification}. Building on this, recent work on prompting generative LLMs (e.g., GPT and Llama) for clinical text de-identification demonstrates that zero and few-shot prompting can effectively de-identify PII (e.g., achieving near-perfect accuracy above 0.95). 

\subsection{Evaluating De-identification}
Regardless of the model, evaluations have remained largely limited to classification metrics (i.e., accuracy, precision, recall, and F1-score). While useful, these metrics fail to capture the full variety of errors which may occur during de-identification. More specifically, false positive errors (i.e., when a token is incorrectly classified as sensitive and removed) can change clinical meaning (e.g., medications, patient history, diagnoses, procedures, and test results). Such a change is worse than a false positive that removes a stop word (e.g., a, the, of). However, traditional classification metrics treat these errors as equal.

\begin{table*}[htbp]
\tiny
  \centering
  \begin{tabular}{llllllllll}
    \hline
    \textbf{Paper} & \textbf{Model} & \textbf{Dataset} & \textbf{Prompting} & \textbf{A} & \textbf{P} & \textbf{R} & \textbf{F1} & \textbf{FNR} & \textbf{FPR} \\
    \hline
    
    \multirow{2}{*}{\citet{altalla2025evaluating}} & GPT-3.5 & \multirow{2}{*}{\parbox[t]{2cm}{\raggedright 100 discharge summaries [PCD]}}

 &\multirow{2}{*}{zero shot} & 0.79 & 0.34 & 0.67 & 0.39 &- & -\\
                             & GPT-4 &   &  & 0.99 & 0.99 & 0.83 & 0.90 & -&- \\
    \hline
    \multirow{14}{*}{\citet{sousa2025large}} &BERT & \multirow{14}{*}{n2c2} & \multirow{4}{*}{fine-tuned} &- & 0.93 & 0.95 & 0.94 & - &-\\
    & ClinicalBERT & & &- &0.84 &	0.85 &	0.84 & - &-\\
    & DistilBERT & & &- &0.90 &	0.92 &	0.91 & - &-\\
    & RoBERTa &  & &- & 0.95 &	0.96 &	0.96& - &-\\
    & FLAN-T5 XXL &   & \multirow{5}{*}{one shot}  & 0.40 & 0.55 & 0.59 & 0.57 &- &- \\
                             & GPT-3.5 Turbo &  &  & 0.45 & 0.65 & 0.59 & 0.62 & -&- \\
                             & GPT-4 &  &  & 0.63 & 0.70 & 0.87 & 0.78 & - & - \\
                             & Llama-3 &  &  & 0.36 & 0.59 & 0.48 & 0.53 & -&- \\
                             & Mistral-7B &  &  & 0.33 & 0.38 & 0.69 & 0.49 &- &- \\
    & FLAN-T5 XXL &  & \multirow{5}{*}{zero shot}  & 0.06 & 0.09& 0.14 & 0.11 &- &- \\
           & GPT-3.5 Turbo &  &  & 0.20 & 0.28 & 0.42 & 0.33 & -&- \\
           & GPT-4 &  &  & 0.41 & 0.48 & 0.75 & 0.58 & - & - \\
           & Llama-3 &  &  & 0.26 & 0.55 & 0.32 & 0.41 & -&- \\
           & Mistral 7B &  &  & 0.14 & 0.15 &  0.58& 0.24 &- &- \\
        
    \hline
    \multirow{8}{*}{\citet{wiest2025deidentifying}} & Llama-3 8B& \multirow{8}{*}{250 clinical letters [PCD]} & \multirow{7}{*}{zero shot}  & 0.98 & 
0.5* &0.99 & -& 0.006 & 0.02 \\
                       
                             & Llama-3 70B &  &  & 0.98 &
0.5* &0.99 & -& 0.01 & 0.02 \\
                             & Llama-2 7B &  &  & 0.97 &
0.5* &0.95 & -& 0.05 & 0.03 \\
                             & Llama-2 70B &  &  & 0.97 & 
0.5* & 0.99 &- & 0.01 & 0.03 \\
                             & Mistral 7B &  &  & 0.98 &
0.5* &0.94 & -& 0.06 & 0.02 \\
                             & Phi-3 Mini &  &  & 0.97 & 
0.3* &0.92 & -& 0.08 & 0.03 \\
                             & LLM-Anonymizer & & &0.98 &
0.8* &0.99 & -&0.01 & 0.02\\
& Presidio & &- &0.71 &
- &0.82 & -&0.17 & 0.29\\
    \hline
    \multirow{5}{*}{\citet{yashwanth2024zero}} & GPT-3.5 & \multirow{5}{*}{i2b2-2014} & \multirow{2}{*}{\parbox[t]{1cm}{(fine-tuned + few shot)}}  & 0.99 & -& -& 0.99 &- &- \\
                             & PaLM &  &  & 0.96 &- &- & 0.95 & -&- \\
                             & GPT-3.5 &  & zero shot & 0.96 &- & -& 0.95 & -&- \\
                             & GPT-4 &  & zero shot & 0.99 & -& -& 0.99 & -&- \\
                             & PaLM &  & zero shot & 0.74 & -& -& 0.73 & -&- \\
    \hline
    \multirow{1}{*}{\citet{chang2024llm}} & Llama-2 70B & institutional safety database[PCD] & zero shot & 0.95 &- &- &- &- &- \\
    \hline
    
%     \multirow{2}{*}{\citet{singh2024generation}} & Llama-3 8B-Instruct &  \multirow{2}{*}{\parbox[t]{2.5cm}{Indian
% Clinical Discharge Summaries[PCD]}} & - & & 0.55 & 0.11 & - & -& -\\
%                              & PI-RoBERTa &   & fine-tuned & & - & - & 0.97 &- &- \\
%     \hline
    \multirow{2}{*}{\citet{langenbach2024automated}\textsuperscript{\textdagger}} & Llama-2 & \multirow{2}{*}{chest reports[PCD]} & fine-tuned & - & 0.99 & 0.84 & 0.89 & - & -\\
                            
                             & NLP &  & rule-based & - & 1 & 0.93 & 0.96 & - & -\\
    \hline
    \multirow{1}{*}{\citet{gunay2024llms}} & GPT-4 & \multirow{1}{*}{i2b2}  & one shot & 0.97& 0.97 & 0.97 & 0.97 & -&- \\
                             % & NER &  & fine-tuned &- & - & - & 0.93 & -&- \\
    
    \hline
    \multirow{6}{*}{\citet{liu2023deid}} 
    & BERT & \multirow{6}{*}{i2b2-2014}  & fine-tuned & 0.80 & -& -&- &- &- \\
    
    & RoBERTa & & fine-tuned & 0.95 &- &- &- &- &- \\
    
    & ClinicalBERT &  & fine-tuned & 0.97 & - &- & -& -& -\\
    
                             & Llama-2 7B & & zero shot & 0.61 &- & -& -& -& -\\
                             & ChatGPT&  & zero shot & 0.93 & -&- &- &- &- \\
                             & GPT-4 &  & zero shot & 0.99 &- & -& -& -&- \\
    
    \hline
  \end{tabular}
  \caption{\label{model-performance}
    Works using generative LLMs for de-identification. Missing values indicate unreported metrics. 
     \textbf{A: Accuracy, P: Precision, R: Recall, F1: F1-score, FNR: False Negative Rate, FPR: False Positive Rate.} *approximated from graph, \textsuperscript{\textdagger} metrics averaged over identifiers 
  }
\end{table*}

This issue is exacerbated by the fact that zero-, one-, or few-shot de-identification by generative LLMs may not always match the input text (i.e., a hallucination unrelated to de-identification). Realizing this issue, \citet{pissarra2024unlocking} proposed six novel metrics beyond classification-based evaluation that explicitly account for clinically significant changes. Two of these metrics, Jaccard Similarity Coefficient (JSC) \citep{jaccard1901etude} and Normalized Softmax Discounted Cumulative Gain (NSDCG), attempt to measure CIR.
JSC uses a BioBERT model trained to predict which ICD-10 codes, codes used to classify medical diseases and conditions, are present in a text. JSC applies this model to the original and de-identified note to measure the overlap in ICD-10 codes predicted from each note. NSDCG takes a similar approach to JSC but does not binarize the outputs, instead comparing the ranking of ICD-10 code predictions (logits) from original and de-identified notes. 

Both of these metrics aim to quantify CIR by comparing the changes in the associated ICD-10 codes. While this is an intuitive approach, in examining our own de-identification task, we observed that these metrics themselves faced limitations. Specifically, they are solely concerned about changes which would change the ICD codes predicted from a text. Although such changes are important to capture, such changes do not constitute the full set of clinically relevant changes (CRC) that may occur. For example, changes in patient history, family status, or employment, which can affect treatment of a patient, will (usually) not change the ICD codes assigned to the note. As such, many CRC are undercounted.

\section{Literature Review}

The sensitive nature of the data and the rapid evolution of LLMs have introduced challenges in benchmarking and reproducibility. Although many studies report high performance, differences in data set choice, prompt design, and evaluation criteria make it difficult to compare findings across the literature. 
To facilitate future research endeavors, we have collated de-identification works that employed generative LLM models (e.g., GPT, Llama) to analyze their performance, the datasets utilized, and to identify potential gaps for future investigation, see Table \ref{model-performance}.

Several studies use a \textit{previously} publicly available datasets such as n2c2 (originally i2b2) \citep{stubbs2015annotating}\footnote{Although the dataset has been made ``Temporarily Unavailable'' for the past two years.}, which provides a standardised benchmark for clinical de-identification \citep{liu2023deid, sousa2025large, yashwanth2024zero}. However, a significant portion of the literature relies on private institutional datasets (e.g., institutional safety reports, clinical letters, discharge summaries, etc.) limiting reproducibility and comparison. 
Some of the existing work uses closed-source models (e.g., GPT-3.5, GPT-4), which, in addition to privacy concerns as data must be sent off-premises, raises concerns regarding reproducibility, in contrast to open-source models (e.g., Llama, Phi). 

Studies also differ in how models are prompted. Some adopt a zero-shot setting, relying purely on the model's pre-trained knowledge without any in-context examples \citep{liu2023deid}, while others use few-shot prompting by providing a few examples to guide the model. \citet{yashwanth2024zero}
fine-tuned GPT-3.5 and PaLM in the i2b2 2014 dataset for clinical de-identification, experimenting with brief and detailed prompt styles.

Most of the metrics used for the evaluation are classification-based (e.g., accuracy, precision, recall, F1 score). LLM-based approaches generally demonstrate high performance, with \citet{liu2023deid} reporting 99\% accuracy for GPT-4. Similarly, \citet{ altalla2025evaluating} showed that GPT-4 outperformed GPT-3.5, achieving 99.25\% precision, 83.18\% recall, 89.73\% F1-score, and 99.11\% accuracy. 

Past work describes LLMs as performing better with few-shot prompting (accuracy and F1-score often in the high 90s) compared to zero-shot attempts (performance varies greatly from as low as 0.39 and high as 0.99), depending on the model and dataset. For example, GPT-3.5 performed poorly (accuracy around 80s) on discharge summaries (private data) \citep{altalla2025evaluating} compared to structured data sets like i2b2 \citep{liu2023deid} (accuracy around mid-90s), highlighting that the reliability of the model can depend heavily on input data and its structure.

In general, we can see that despite the tremendous amount of effort undertaken by researchers, the inconsistencies regarding the dataset used and the metrics reported severely limit our ability to compare de-identification approaches, and thus the utility of the work to readers and real-world NLP users.

\section{Datasets}
We conducted experiments on two different English-language clinical datasets: a large public dataset (MIMIC-III) \citep{johnson2016mimic} and a smaller private dataset of %AHS 
rheumatology referral notes. The appendix Table \ref{tab:data_descriptive} presents some descriptive statistics of the two datasets.

\subsection{MIMIC}

We randomly sampled 2000 de-identified discharge summaries to provide some comparison to \citet{pissarra2024unlocking}, who also sampled discharge summaries. Since the notes are already de-identified, we used the Faker library \citep{Faraglia_Faker} to randomly replace the placeholders based on their relative tags, illustrated in Appendix Figure \ref{fig:faker-library}.

\subsection{Private Clinical Dataset (AHS)}
Our private dataset contains 204 referral letters from physicians to rheumatologists documenting a variety of clinical scenarios, physical exam findings, medication lists, lab and imaging results, and other clinical notes. The collection and use of this dataset was approved by the University of Alberta's REB (\#Pro00141020). All of these notes were stored in PDF format, including a mix of digitally typed notes, scanned documents, and handwritten letters. We performed OCR (Optical Character Recognition) with the Doctr library \citep{doctr2021}. A computer scientist and a physician manually removed all sensitive tokens, excluding provider/clinic information and years in the date of birth. To enable de-identification, we replaced all removed identifiers using the Faker library, injecting noise (e.g., character and number swaps) into a subset of the replaced tokens to match the original OCR results.

\section{Models}
We tested multiple models: generative LLMs (Llama 3.3), ClinicalBERT, two established open source toolkits (Deidentify and Presidio) and a combined approach of ClinicalBERT and Presidio.

\begin{figure*}[h]
\centering
\includegraphics[width=0.85\textwidth]{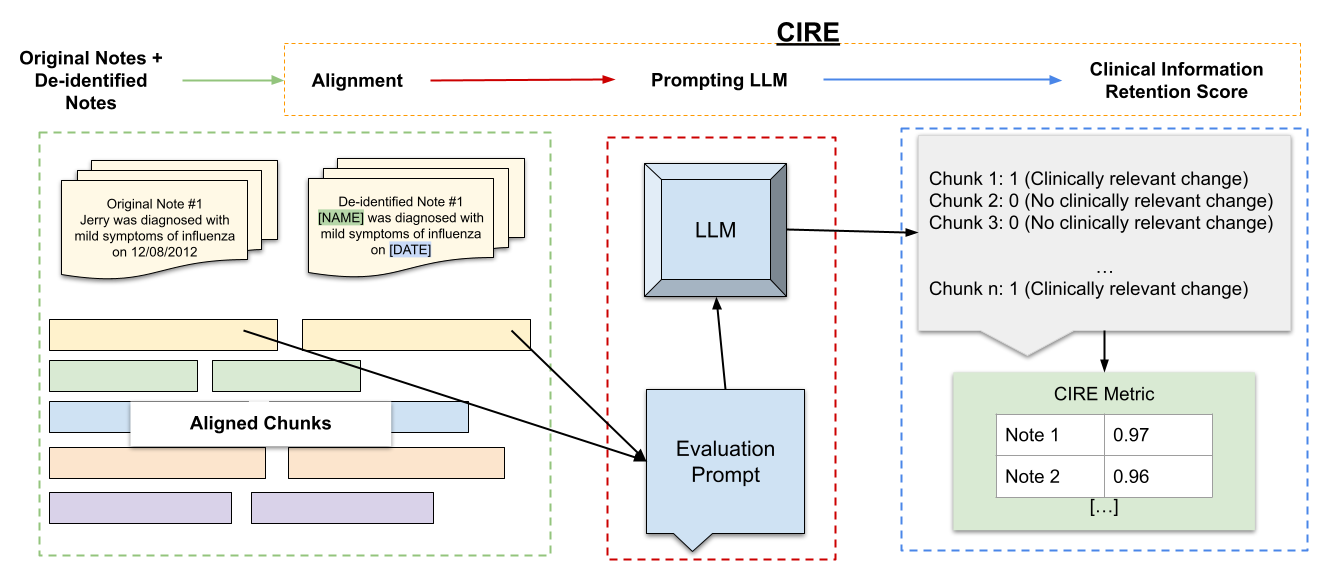}
\caption{Illustration of the novel LLM-based Clinical Information Retention Evaluation (CIRE) approach. \label{fig:LCIRE_fig}}
\end{figure*}

\subsection{Llama 3.3}
Llama 3.3 (70B parameters, 16 bits per parameter) was used with prompt tuning on 10\% of the private dataset. We used an iterative trial-and-error approach to prompt design. Our prompting strategy followed a one-shot format, where a single annotated example was provided to guide the model. The final prompt is presented in Appendix Figure \ref{fig:llama_deid_prompt}. During prompt tuning, multiple prompts were tested on the AHS dataset, and the prompt that achieved the highest recall was selected and then adapted for the MIMIC-III dataset to target its specific sensitive tokens. The results of the performance of prompt tuning are presented in the Appendix Table \ref{tab:llama_deid_train}.

\subsection{ClinicalBERT}
ClinicalBERT \citep{alsentzer2019publicly}, pre-trained in MIMIC-III and fine-tuned in i2b2 / n2c2, was used \citep{obi_deid_bert_i2b2}. To accommodate long clinical notes, we split the text into 512-token chunks, processed each chunk individually, and then reconstructed the full de-identified note. 

\subsection{Deidentify}
The Deidentify Python library \citep{Trienes:2020:CRF} combines pre-trained neural sequence-tagging models with rule-based patterns (e.g., regular expressions for dates and phone numbers).

\subsection{Presidio}
Microsoft Presidio, an open-source framework, detects PII (named: entities) using NER (named entity recognition) techniques.

\subsection{Bert-Presidio}
In the Bert-Presidio pipeline, we applied a modified ClinicalBERT model (described below) and then ran the text through the modified presidio with some built-in and custom entity sets.

\subsection{Modified Versions}
For all ``off-the-shelf'' algorithms, we also evaluate a modified version (e.g., ``Modified ClinicalBERT'', ``Modified De-identify''). Given that MIMIC is an American dataset, to enable a fair comparison, we needed to add regex patterns to catch non-US PII (e.g., addresses, postal codes and patient health number) in addition to keeping age and only the birth year in dates or birth. Likewise, some of the pre-selected identifiers did not apply to our dataset. Therefore, for ``Modified Presidio'' we only anonymized selected entities such as `Person' and `Phone number'. 

\begin{table*}[ht]
\small
  \centering
  \begin{tabular}{lllllllllll}
    \hline
    \textbf{Dataset} & \textbf{Model} & \textbf{TP} & \textbf{TN} & \textbf{FP} & \textbf{FN} & \textbf{A} & \textbf{R} & \textbf{P} & \textbf{F1} &  \\
    \hline
    \multirow{8}{*}{AHS} & Presidio & 5508 & 120245 & 26544&1772 & 0.82 & 0.76& 0.17 &  0.28

\\
    & Modified Presidio & 5039 &133312 &13477 & 2241& 0.90& 0.69& 0.27&0.39 & \\
    & ClinicalBERT & 6806  & 119392&27397 &474 &0.82 &0.93 &0.20 &0.33 & \\
    & Modified ClinicalBERT & 6974 & 118804& 27985& 306& 0.82&0.96 & 0.20&0.33 & \\
    & ClinicalBERT + Presidio & 6996&117180 & 29609&284 &0.81 & 0.96& 0.19& 0.32 \\
    & Deidentify &5689 & 129339&17450 &1591 &0.88 &0.78 & 0.25& 0.37 \\
    & Modified Deidentify &6117 &128304 &18485 & 1163& 0.87&0.84 &0.25 & 0.38 \\
    & Llama-3 & 7214&144385 &2404 &66 &\textbf{0.98} &\textbf{0.99} &\textbf{0.75} & \textbf{0.85} \\
    \hline
    \multirow{8}{*}{MIMIC-III} &Presidio &113716 &2688543 &216510 &28252 &0.92 &0.80 & 0.34& 0.48 \\
    & Modified Presidio &106739  & 2881778&21988 &34674 &0.98 &0.75 &0.83 & 0.79& \\
    & ClinicalBERT & 134831 & 2888521&16532& 7137&\textbf{0.99}& \textbf{0.95}& \textbf{0.89}&\textbf{0.92} & \\
    & Modified ClinicalBERT & 134959 &2885440& 19613&7009 &\textbf{0.99} &\textbf{0.95} &0.87 &0.91& \\
    & ClinicalBERT + Presidio &133677 & 2861053& 44000&8291 &0.98 &0.94&0.75& 0.84 \\
    & Deidentify &118047 &2873672 & 31381&23921 & 0.98&0.83 &0.79 & 0.81 \\
    & Modified Deidentify & 118096& 2871166&33887 &23872 & 0.98& 0.83&0.78 & 0.80 \\
    & Llama-3 &133107 &2854895&50158&8861 &0.98 & 0.94&0.73 &0.82  \\
    \hline
  \end{tabular}
  \caption{Standard Classification Metrics
  \textbf{TP: True Positive, TN: True Negative, FP: False Positive, FN: False Negative, A: Accuracy, P: Precision, R: Recall, F1: F1-score }}
  \label{tab:standard-evaluation}
\end{table*}

\section{Evaluation Metrics}
\subsection{Standard Classification Metrics}
To start, we used existing evaluation metrics (accuracy, precision, recall, and F1 score). Accuracy is the ratio of correctly predicted tokens (sensitive and non-sensitive) to the total number of tokens.
% \begin{equation}
% Accuracy = \frac{TP + TN}{TP + TN + FP + FN}
% \end{equation}
Precision is the proportion of tokens predicted as personally identifying that are actually personally identifying.
% \begin{equation}
% Precision = \frac{TP}{TP + FP}
% \end{equation}
Recall evaluates the model's ability to detect all personally identifiable tokens in the data set.
% \begin{equation}
% Recall = \frac{TP}{TP + FN}
% \end{equation}
F1-score is the harmonic mean of precision and recall.
% \begin{equation}
% F1~Score = \frac{2  \times Precision  \times Recall}{Precision + Recall}
% \end{equation}

\subsection{Evaluating Clinically Significant Changes}
While standard classification metrics are essential for evaluating the identification of sensitive entities, they do not fully reflect the "cost" or impact of different error types. Specifically, such metrics do not differentiate between the clinical relevance of falsely removed tokens and do not consider the downstream impact of de-identification on the clinical utility of the text. In this section, we present two metrics, one existing and one novel, to evaluate how much LLM result in CRC.

\subsubsection{Jaccard Similarity Coefficient (JSC)}
JSC, developed by \citet{pissarra2024unlocking}, uses a trained BERT model to predict ICD-10 codes from text to measure CIR. This approach measures changes in ICD codes predicted from a note before and after de-identification as a proxy for CIR \citep{pissarra2024unlocking}. Changes in the ICD-10 codes after de-identification indicates that clinically relevant information has been altered (as they fairly assume that removing PII (e.g., names) should not affect ICD-10 codes prediction). We used the JSC baseline settings as specified by \citet{pissarra2024unlocking}.

\subsubsection{LLM-based Clinical Information Retention Evaluation (CIRE)}
Modifications that change the predicted ICD codes are only a subset of all possible CRC. For example, removal of patient history, test result numbers, or procedures can affect patient treatment (thereby counting as clinically relevant) without changing the ICD code predicted from the note.  Due to the above limitations, we argue that JSC alone is insufficient for evaluating whether clinically relevant information has been preserved through de-identification. To address this gap, we propose LLM-based Clinical Information Retention Evaluation (CIRE), a method that directly assesses clinically relevant semantic changes using a sentence-level LLM-based prediction.% a sentence-level alignment-based method that directly assesses semantic changes in the clinical notes using LLM-based evaluation.

To assess whether the de-identification process inadvertently changed clinically relevant information, CIRE works as follows (illustrated in Figure \ref{fig:LCIRE_fig}): First, notes before and after de-identification are split into sentences that are then aligned using the string2string library \citep{suzgun2023string2string}. Once aligned, the token sequences from both sentences are grouped into fixed-length chunks (20 tokens per chunk). Our choice to use fixed-length chunking rather than sentence-level splitting was primarily motivated by OCR-related noise in the dataset and supported by experimental results; further details are provided in Appendix~\ref{sec:chunking-vs-sentence}. Paired chunks are presented to a Llama 3 model which is prompted to classify ``whether the deidentified sentence chunk has altered any clinically meaningful information'', complete prompt presented in Appendix Figure \ref{fig:evaluation_prompt}. Then each sentence is assigned a 1 or 0 for clinically relevant alteration. For each note, the per-sentence classifications are then averaged to produce a metric of CIR. The pseudocode for CIRE is presented in the Appendix Algorithm \ref{alg:alignment-chunking}, and the results of the training set can be found in Appendix \ref{sec:LCIRE Development}.

\begin{figure}[h]
\includegraphics[width=\columnwidth]{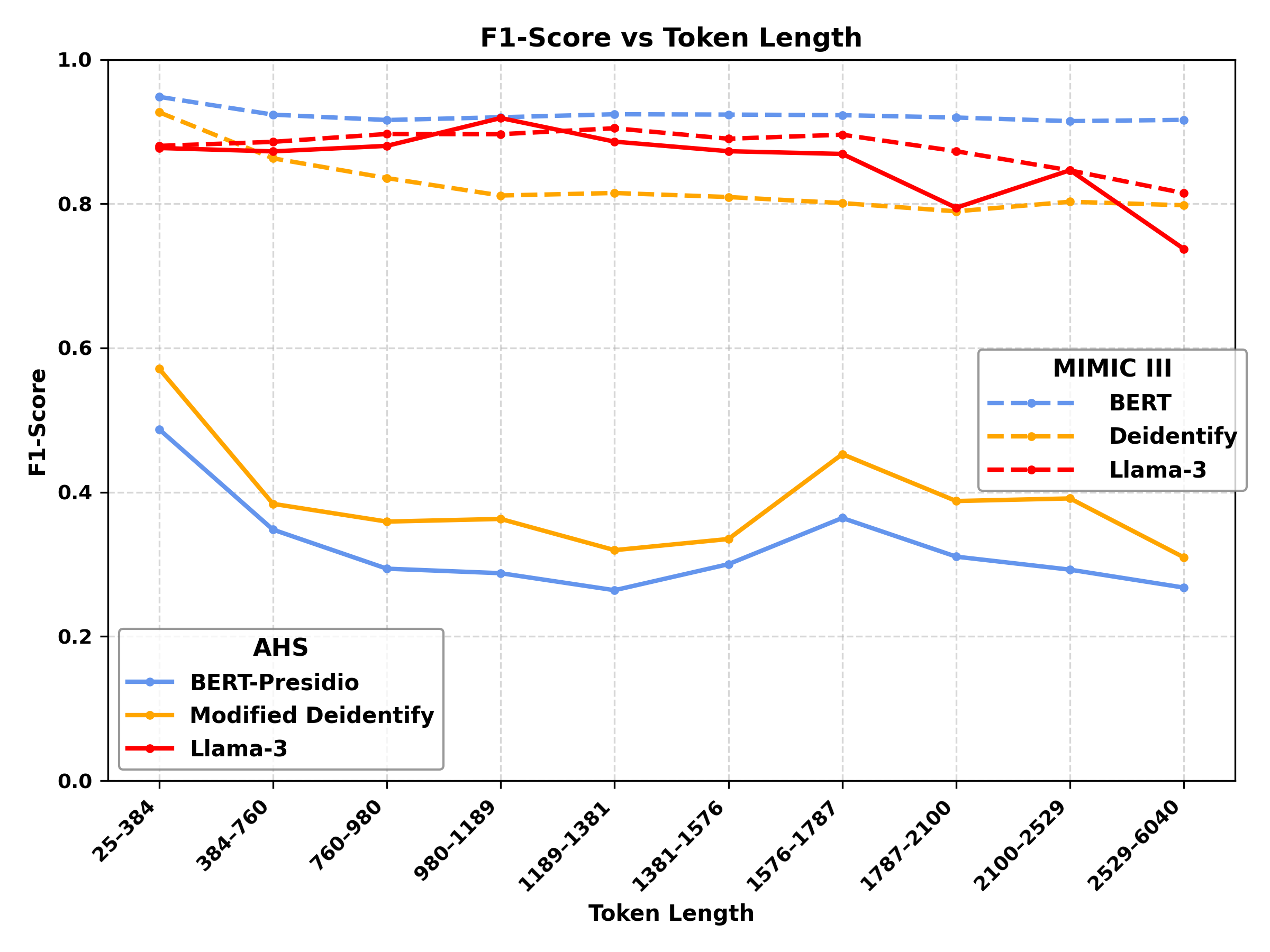}
\caption{F1-Score of three de-identification models on AHS and MIMIC binned by the number of tokens in the clinical texts. 6,040 tokens corresponds to approximately 4,500 words.
\label{fig:perf_by_length}}
\end{figure}

\section{Experiments and Results}
\subsection{How do de-identification models perform on standard evaluation metrics? How does document length affect performance?}
We evaluated models using traditional classification metrics (accuracy, precision, recall, F1 score) in AHS and MIMIC-III and present the results in Table \ref{tab:standard-evaluation}. Llama-3.3 was the highest performer on AHS (near perfect recall and highest precision and F1 score). For MIMIC-III, while all models achieved accuracy in the high 90s, there was a large variation in the other metrics. This is because the accuracy is not robust to class imbalance, and PII is relatively rare. This highlights the inappropriateness of simply reporting accuracy as done in two previous works. However, we observe that BERT variants achieved relatively higher precision compared to Llama, which could be due to the fact that the ClinicalBERT model we used was pre-trained on the MIMIC dataset (so data leakage may be present). %These results highlight that model performance can vary notably across datasets and that pre-training approaches can still provide benefits not met by the generalization of LLMs.
Further analyses of schema variants are provided in Appendix~\ref{sec:deid-variations}, including evaluations where provider identifiers are considered PII and where sub-token handling differs.
Although generative LLMs are easier to use `out of the box' (i.e., they do not require fine-tuning and are more generalizable), they have been reported to under-perform as input length increases \citep{wang2024benchmarking}. 
Binning the de-identification evaluation of Deidentify, BERT and Llama models by token lengths for both AHS and MIMIC-III, we observe that this observation is not limited solely to generative LLMs. The F1 score of the three models decreases between 5-25\% as the length of the text increases, Figure \ref{fig:perf_by_length}.

\subsection{How successful are de-identification models at clinical information retention?}
\begin{table}[t]
  \small
  \centering
  \begin{tabular}{lllll}
    \hline
    \textbf{Dataset} & \textbf{Model} & \textbf{CRC(\%)} & \textbf{Low} & \textbf{High}\\
    \hline
    \multirow{3}{*}{AHS} & M. Deidentify & 22 (4) & 7& 15\\
    % & BERT & &  \\
    & CBERT + Pres & 14 (3)&  4& 10\\
    & Llama-3.3 &89 (18) & 22& 67\\
    \hline
    \multirow{3}{*}{MIMIC-III} & Deidentify & 444 (89) & 57&387 \\
    & CBERT &263 (53) &45 & 218\\
    % & BERT + Presidio & & & & \\
    & Llama-3.3 &252 (50) &57& 195\\
    \hline
  \end{tabular}
  \caption{Number of clinically relevant changes (CRC) caused by three de-identification models on two different datasets. For each model, 500 false positive cases were manually annotated. Then each CRC was labelled with a clinical severity of low or high. M. Deidentify: Modified De-identify. CBERT: ClinicalBERT. Pres: Presidio}
  \label{tab:clinical-changes-evaluation}
\end{table}
A high rate of false positives in Table \ref{tab:standard-evaluation} indicates a tendency toward over-redaction, with all models redacting much more content than necessary. This raises concerns about the loss of meaningful clinical information. This is further compounded by the fact that LLMs may make changes unrelated to the task of de-identification (current approaches, including our own, do not make predictions at the token level rather processing the whole note at once). To better understand the impact of false positives, for each algorithm, we randomly sampled 500 changes that were incorrectly redacted by a subset of the evaluated models (i.e., false positive). Each change was manually reviewed by a computer scientist and a physician to determine whether its removal resulted in CRC, categorization presented in Table \ref{tab:errors-classification}. To ensure consistency across annotators, we provided detailed written guidelines, which are included in Appendix~\ref{sec:annotators}.  The physicians then assessed each CRC to determine if the clinical impact caused by the removal of the information was high or low. The results are presented in Table \ref{tab:clinical-changes-evaluation}, and a description of high and low clinical impact is presented in the Appendix Section \ref{sec:high_and_low}.

\begin{table*}[!htbp]
  \centering
  \begin{tabular}{l l cccc}
    \hline
    \textbf{Dataset} & \textbf{Model} & \multicolumn{2}{c}{\textbf{Pearson}} & \multicolumn{2}{c}{\textbf{Spearman}} \\
    \cline{3-4} \cline{5-6}
                    &                & CIR-JSC & CIR-CIRE & CIR-JSC & CIR-CIRE \\
    \hline
    \multirow{4}{*}{AHS}         & Modified Deidentify     & 0.43	 & 0.78 & 0.43 &	0.51 \\
          & ClinicalBERT + Presidio & -0.51	& 0.67 & -0.25	& 0.57 \\
          & Llama-3.3                 & 0.43	& 0.73 & 0.6& 0.48 \\
          & All                     & -0.34 &	0.78 & -0.20 &	0.59 \\
    \hline
    \multirow{4}{*}{MIMIC-III}         & Deidentify     &  -0.03	& 0.86 & -0.21	& 0.88 \\
          & ClinicalBERT + Presidio & 0.19 &	0.85 & 0.41 &	0.84 \\
          & Llama-3.3                 & 0.09 &	0.33 & 0.07 &	0.40 \\
          & All                     & 0.04	& 0.40 & 0.17 &	0.41 \\
    \hline
  \end{tabular}
  \caption{Correlation metrics subdivided into CIR-JSC and CIR-CIRE for Pearson and Spearman. M. Deidentify: Modified De-identify. CBERT: ClinicalBERT. Pres: Presidio.}
  \label{tab:correlation}
\end{table*}

We can see that a substantial subset of these redacted tokens was found to affect clinically relevant information, with Llama-3.3 showing the highest number of clinical redactions (89 out of 500) on AHS and Deidentify with the highest count (444 out of 500) for MIMIC-III dataset. Unfortunately, even for the best performing models, most (70-80\%) false positives caused substantial changes in clinical information that doctors considered high.

\subsection{Can automated metrics accurately measure CIR?}
%Preserving the clinical utility of text is a critical objective in de-identification. Yet, 
As highlighted in the previous section, all models introduce CRC that physicians consider to be highly impactful. To assist with the evaluation (and improvement) of de-identification approaches, we investigate whether CIR can be assessed automatically. To do this, we evaluated the JSC and our novel metric (CIRE). 

This experiment leveraged 40 notes (30 AHS, 10 MIMIC-III) that were manually labeled at the sentence level (6720 annotated sentences) comparing pre- and post-de-identification for CRC. We used 20 full-text AHS text as our prompt-tuning set. We evaluated the performance of CIRE to detected CRC on the remaining full texts (10 AHS and 10 MIMIC-III).

On this subset of data, we evaluated CIR after de-identification using both the existing metric JSC and our novel metric CIRE. We computed Pearson and Spearman correlations between the results of CIRE and JSC and the existing ground truth (ratio of total clinically unchanged sentence pairs to the total sentence pairs across each document), Table \ref{tab:correlation}. Although the precision and recall of CIRE could be substantially improved (Table \ref{tab:lcire-evaluation}), our analysis revealed a strong Pearson correlation between CIRE and manually labeled CIR (ground truth) that was much stronger than with JSC (Table \ref{tab:correlation}). This demonstrates that JSC is indeed insufficient as a metric for assessing CIR while, at the same time, demonstrating the potential of our approach. We are excited for the community to work to further improve performance.

When we use CIRE to evaluate the clinical retention of Llama-3.3, ClinicalBERT, and Deidentify on the entire AHS dataset, we observe that all models score in the mid-90s with ClinicalBERT performing the best with the average amount of clinical retention at 95\%, Appendix \ref{sec:lcire_application}, Table \ref{tab:manual-evaluation}. \\
While our main evaluation used Llama-3.3 as the reference LLM, we verified that CIRE remains stable for other reference LLMs. As detailed in Appendix~\ref{sec:cire-robustness}, results with Qwen-72B, Qwen-32B, and Llama-3.1-8B show consistent relative rankings across models.

\section{Conclusion}
In this work, we started by exploring the current evaluation methodologies of papers performing de-identification using generative LLMs. We found that there was low consistency in the datasets used or the metrics reported -- a problem that reduces the utility of the conducted research.

We then evaluated the performance of a wide variety of models on two different datasets. We demonstrated that despite what is claimed by prior work, there is a wide variation in performance and non-generative models can sometimes beat generative LLMs in de-identification. 

Next, we highlighted that the many false positives caused by the process of de-identification actually remove clinical relevant information and that most of these removals were deemed, by physicians, to be significant (i.e., high severity). We then demonstrated that existing metrics aiming to quantify the effect of de-identification with respect to CIR were not sufficient, and are fundamentally flawed in their approach. 

We conclude the paper by introducing a new metric, CIRE. CIRE is more strongly correlated with manual counts of CRC and presents a more reliable assessment of CIR. We applied it on both datasets and observed that all models both traditional and new seem to retain clinical information at nearly the same rate (differences in performance as measured by CIRE were not substantial).

\subsection{Future Work}

This paper has uncovered several avenues for future work. Those working on de-identification need to tackle the problem of inappropriate clinical information removal possibly by developing a multi-stage de-identification process where proposed removals are assessed for clinical relevance.

Those interested in evaluation methodologies can work to iterate and improve CIRE. We believe that there is much that exists in the subfield of paraphrase detection that can be used for this task. We are currently creating a dataset to enable a shared task to bring these two communities together.

\section*{Limitations}
There are limitations to our work.
Because of the sensitive nature of the data, our experiments were limited to two datasets (AHS and MIMIC-III). This has various impacts. First, the models, prompts, and rules we have tuned here are unlikely to be generalized to other hospitals, clinical note structures, or narrative styles without additional tuning. Likewise, the performance patterns observed (e.g., ranking of models) may change when applied to different datasets of notes. Third, many of the pre-trained models have been applied to MIMIC dataset (if not fine-tuned on some subset of the MIMIC dataset given its rare status as one of the only few publicly available clinical datasets). This affects the generalizability of our findings. However, since the focus of our work was on the retention of clinical information (or lack thereof), we do not believe that such contamination necessarily invalidates our findings. If anything, we are overestimating the retention of such models (since overall performance would be lower had there been no contamination). 

Another limitation is the presence of tagging errors in the MIMIC dataset. First, some errors involve misclassification of PII types (e.g., names labeled as addresses). While this reduces the realism of replacements generated with Faker, the affected tokens remain PII and are therefore still valid for de-identification. Second, the rule-based system occasionally tags normal tokens as PII; in our manual validation of 2000 disagreements, we observed 90 such cases. In these instances, our Faker approach still produces plausible PII for removal, preserving the validity of evaluation. Third, the rules occasionally miss true PII; we identified only four such examples of 2000 disagreements that we checked, all involving single initials. Given their rarity, these errors are unlikely to substantially affect our findings.

Finally, our work is restricted to English, and the findings may not generalize to other languages or multilingual clinical contexts.

Despite limiting the analysis to these two datasets, running Llama-3.3 on lengthy clinical notes requires substantial GPU memory and processing time, which presented a significant resource challenge. We estimated that the total number of GPU hours used for the experiments conducted in this paper is approximately 1550, Appendix \ref{sec:compute}. As a result, following previous work \citep{pissarra2024unlocking}, we were only able to test on a subset of our data set for de-identification and evaluation (instead of the full MIMIC-III dataset).

Likewise, our evaluation of the CIR of these models was manual, requiring expert physician annotators, which limited the number of instances that could be annotated (and thus the generalizability of our findings).

The application of de-identification is generally low-risk. In fact, this type of application serves to reduce the risk associated with other work. So, in our opinion, the primary effect of this kind of research is positive. However, there may be negative externalities to our work. For example, our proposed metric, which requires processing via a generative LLM, increases compute costs which may also have negative effects on the environment. Likewise, the development of these approaches requires substantial time from physicians which may theoretically be spent providing care to patients. Our work is limited in that we do not account for such externalities.

\section*{Acknowledgments}
This research was funded by Arthritis Society Canada (and partners as applicable) (grant \# SOG-24-0377). Carrie Ye is supported by a CRAF (CIORA)-Arthritis Society Canada New Clinical investigator Award (award \#Cl-24-0013). Ross Mitchell is the Alberta Health Services Chair in Artificial Intelligence in Health and is supported by CIFAR, University Hospital Foundation, Amii, and the Canadian Foundation for Innovation. Mohamed Abdalla is supported by a CIFAR AI chair and an Amii grant.

\bibliography{custom}

\newpage

\appendix
\section{Dataset Description}

\begin{table}[h]
\begin{tabular}{l|cc} & \textbf{AHS} & \textbf{MIMIC-III} \\ \hline
Number of texts & 204 & 2000 \\ \hline
Token Length & & \\
\hspace{3mm}min  & 141 & 26 \\
\hspace{3mm}mean & 755 & 1524 \\
\hspace{3mm}max & 5436 & 6040  \\ \hline
\begin{tabular}[c]{@{}l@{}}Number of Sensitive\\ Tokens\end{tabular} & & \\
\hspace{3mm}min & 8 & 7 \\
\hspace{3mm}mean & 35 & 70 \\
\hspace{3mm}max & 201 & 293 
\end{tabular}
\caption{Descriptive statistics of datasets used in this paper. \label{tab:data_descriptive}}
\end{table}

\section{Faker Library}
\begin{figure}[h]
\includegraphics[width=\columnwidth]{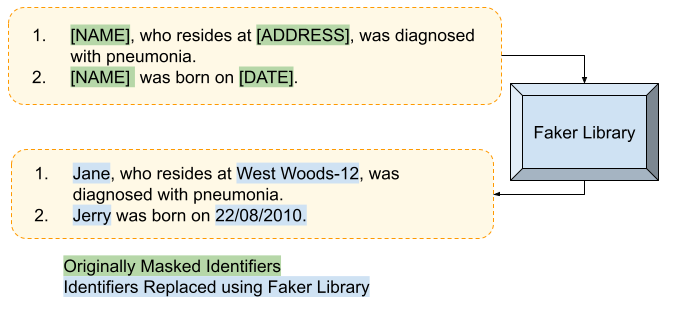}
\caption{Demonstration of Faker Library.
\label{fig:faker-library}}
\end{figure}

\section{Prompts}
\begin{figure*}[h!]
  \centering
  \includegraphics[width=0.9\textwidth]{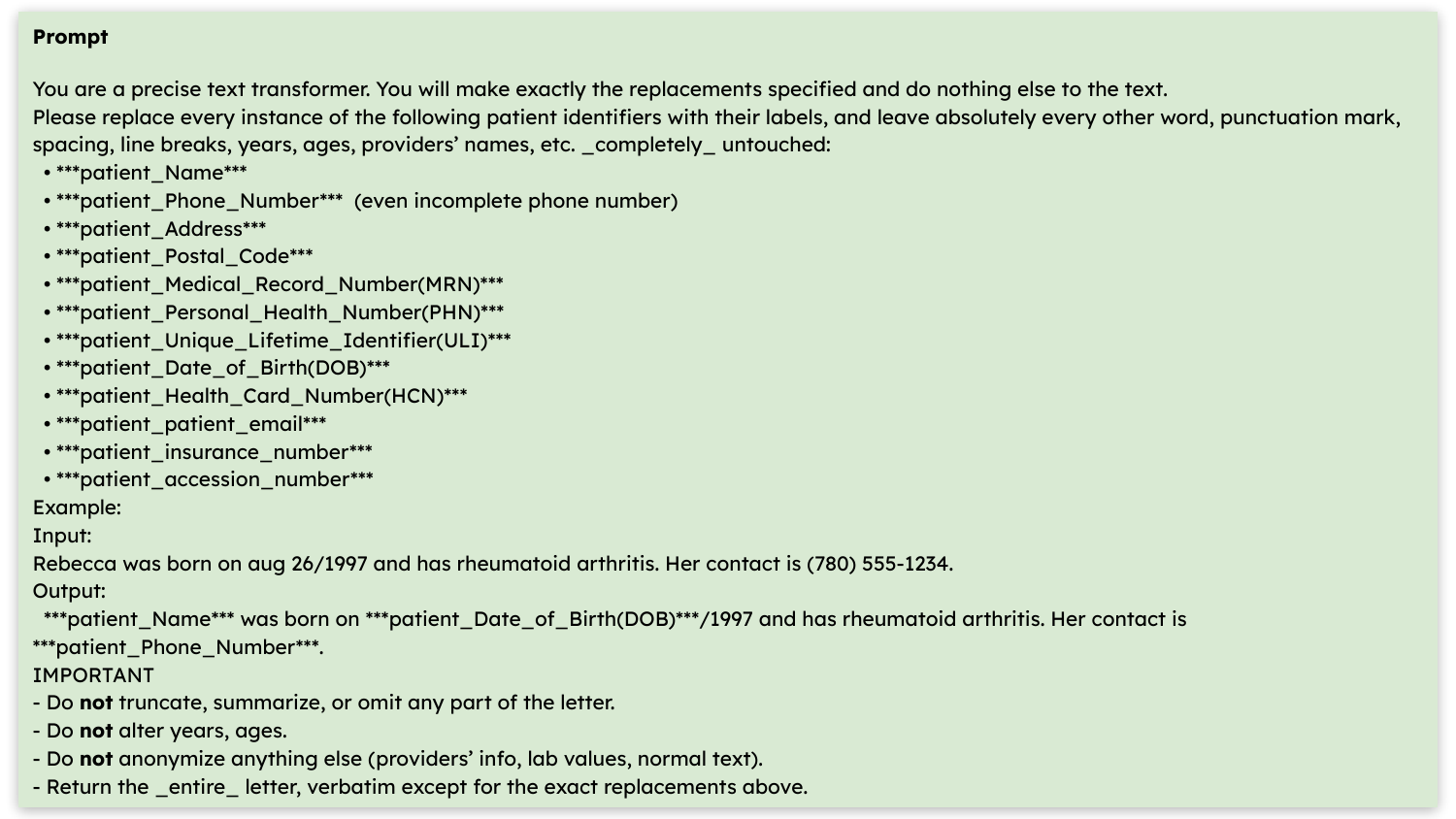}
  \caption{Final prompt used for de-identification using Llama-3.3.}
  \label{fig:llama_deid_prompt}
\end{figure*}

Figure \ref{fig:llama_deid_prompt} presents the final prompt used for the experiments described in the paper.

\begin{figure*}[h!]
\centering
  \includegraphics[width=0.9\textwidth]{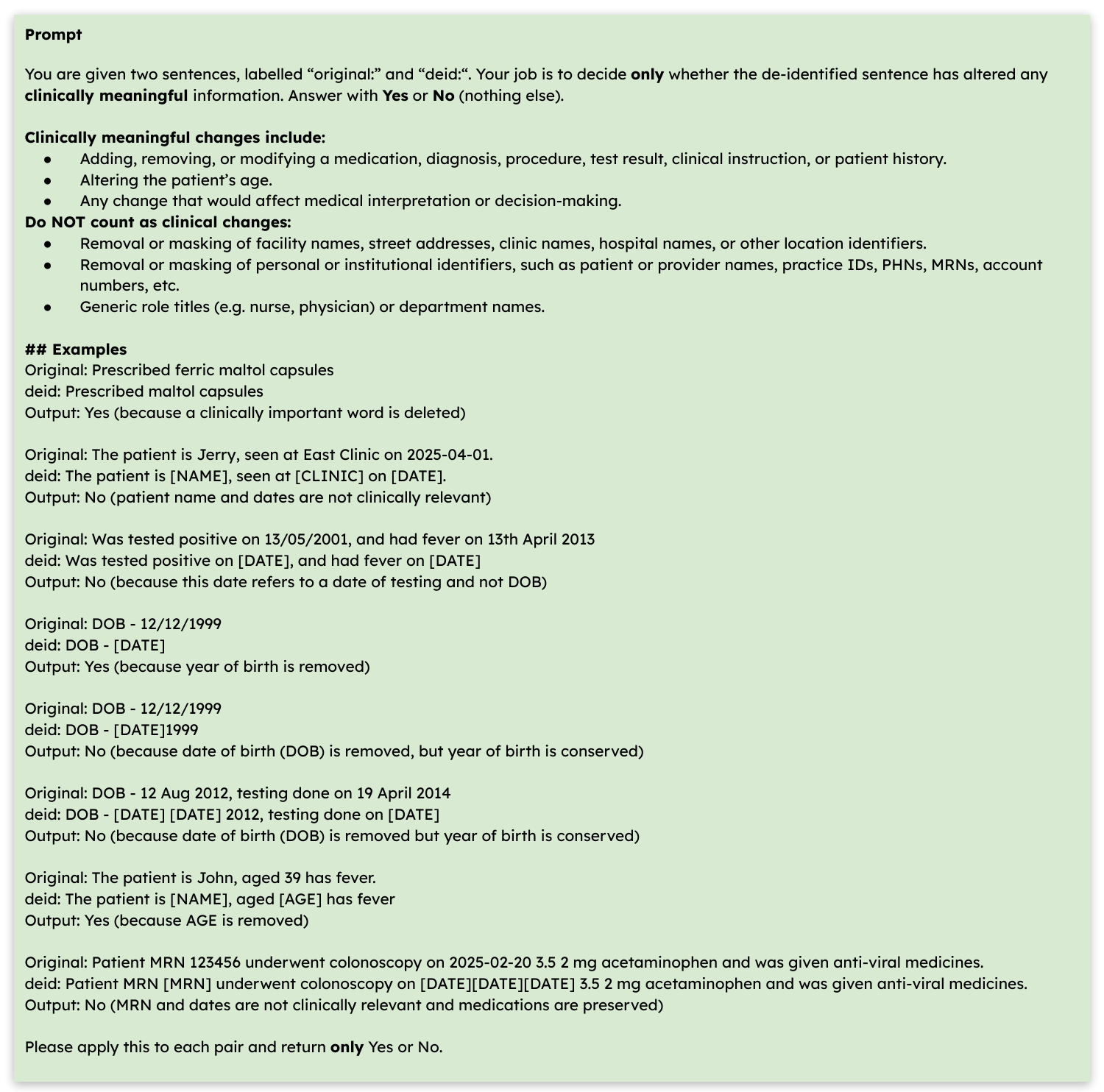}
  \caption{Prompt used for CIR evaluation with Llama-3.3}
  \label{fig:evaluation_prompt}
\end{figure*}

\noindent Figure \ref{fig:evaluation_prompt} presents the final prompt used for the CIR evaluation using Llama-3.3 (i.e., the CIRE metric).

\section{Llama-3.3 De-identification Prompt Tuning}
The performance of Llama-3 after prompt-tuning on the training set is presented in Table \ref{tab:llama_deid_train}.

\begin{table}[h]
\begin{tabular}{l|c}
          & \textbf{Llama-3.3} \\ \hline
TP        & 1020             \\
TN        & 23841            \\
FP        & 610              \\
FN        & 0                \\ \hline
Precision & 0.63             \\
Recall    & 1.00             \\
F1-Score  & 0.77             \\ \hline
\end{tabular}
\caption{Performance of Llama-3.3 after prompt-tuning on the training set. \label{tab:llama_deid_train}}
\end{table}

\newpage
\section{LLM-based Clinical Information Retention Evaluation (CIRE)}
\begin{algorithm}[h]
\small
\caption{Pseudo-code for LLM-based Clinical Information Retention Evaluation (CIRE)}
\label{alg:alignment-chunking}
\begin{algorithmic}[1]
\State \textbf{Initialize:} \texttt{NW} $\gets$ \texttt{NeedlemanWunsch()}
\Function{Chunk}{sequence, size}
    \State tokens $\gets$ \Call{Split}{sequence, `` | ''}
    \State \Return \Call{SplitIntoSublists}{tokens, size}
\EndFunction
\Function{llm}{chunk$_1$, chunk$_2$}
    \State yes/no $\gets$ Llama(prompt,chunk$_1$, chunk$_2$)
    \State \Return Yes or No
\EndFunction

\ForAll{summary}
    \State seq$_1$ $\gets$ \Call{Tokenize}{original summary}
    \State seq$_2$ $\gets$ \Call{Tokenize}{deid summary}
    \State (aligned$_1$, aligned$_2$) $\gets$ \Call{nw.get\_alignment}{seq$_1$, seq$_2$}
    \State chunks$_1$ $\gets$ \Call{Chunk}{aligned$_1$, 20}
    \State chunks$_2$ $\gets$ \Call{Chunk}{aligned$_2$, 20}
    \For{$(c_1, c_2)$ \textbf{in} \Call{zip}{chunks$_1$, chunks$_2$}}
        \State llm input $\gets$ ``original: '' $+$ \Call{Join}{$c_1$, `` ''} $+$
        \State \hspace{1.3cm} ``\textbackslash ndeid: '' $+$ \Call{Join}{$c_2$, `` ''}
        \State decision $\gets$ \Call{llm}{$c_1$, $c_2$}
    \EndFor
\EndFor
\end{algorithmic}
\end{algorithm}

\section{Severity of Clinical Changes}
\label{sec:high_and_low}
For each CRC caused by de-identification, we asked physicians to select if the change was high or low severity. A change was considered high severity if it resulted in loss of clinically meaningful information that could possibly affect diagnosis or management. Changes which would not affect diagnosis or management were annotated as low impact/severity.

\newpage
\section{CIRE Development}
\label{sec:LCIRE Development}
Table \ref{tab:lcire-evaluation} presents the performance of the CIRE pipeline on the manually annotated training set of AHS (20 texts). Table \ref{tab:lcire-in-depth-evaluation} presents the true and false positives and negatives.

\begin{table}[!htbp]
  \small
  \setlength{\tabcolsep}{2pt}
  \centering
  \begin{tabular}{lllll}
    \hline
    \textbf{Dataset} & \textbf{Model}  & \textbf{Accuracy} & \textbf{Precision} & \textbf{Recall}  \\
    \hline
    \multirow{3}{*}{AHS} & M. Deidentify & 0.94 &	0.86 &	0.57 \\
    % & BERT & &  \\
    & CBERT + Presidio & 0.96 &	0.87 &	0.54 \\
    & Llama-3.3 & 0.98 &	1	& 0.71\\
    \hline
    \multirow{3}{*}{MIMIC-III} & Deidentify  & 0.93 & 0.73 & 0.61 \\
    & CBERT & 0.95 & 0.59 & 0.86 \\
    % & BERT + Presidio & & & & \\
    & Llama-3.3 & 0.88 & 0.11 & 0.8\\
    \hline
  \end{tabular}
  \caption{Performance of the CIRE metric as measured by manual annotation on the AHS training set (20 notes).  M. Deidentify: Modified Deidentify; CBERT: ClinicalBERT}
  \label{tab:lcire-evaluation}
\end{table}

\begin{table}[!htbp]
\tiny
  \centering
  \begin{tabular}{llllll}
    \hline
    \textbf{Dataset} & \textbf{Model} & \textbf{TP} & \textbf{TN} & \textbf{FP} & \textbf{FN} \\
    \hline
    \multirow{3}{*}{AHS} 
      & Modified Deidentify      &    24	& 359	& 4	& 18    \\
      & ClinicalBERT + Presidio  &     14	& 377	& 2	& 12    \\
      & Llama-3.3                 &     15	& 385	& 0	 &6     \\
    \hline
    \multirow{3}{*}{MIMIC-III} 
      & Deidentify              &     36 & 479 & 13 & 23     \\
      & ClinicalBERT            &     32	& 491	 & 22	& 5     \\
      & Llama-3.3                 &    8 & 478 & 64 &2   \\
    \hline
  \end{tabular}
  \caption{Results of CIRE on the training dataset.}
  \label{tab:lcire-in-depth-evaluation}
\end{table}

\section{CIRE Application}
\label{sec:lcire_application}
When applied to the full dataset, we can see that all models have similar CIRE scores with an average number of 5\% loss in clinical information.

\begin{table}[!htbp]
\small
  \centering
  \begin{tabular}{llll}
    \hline
    \textbf{Dataset} & \textbf{Model}  & \textbf{JSC} & \textbf{CIRE}  \\
    \hline
    \multirow{3}{*}{AHS} & Modified Deidentify &67.53& 0.90 \\
    % & BERT & &  \\
    & ClinicalBERT + Presidio &  54.56 & 0.95\\
    & Llama-3.3 &71.26 & 0.93\\
    \hline
    % \multirow{3}{*}{MIMIC-III} & Deidentify  & 72.80 & \\
    % & ClinicalBERT & 69.56 & \\
    % % & BERT + Presidio & & & & \\
    % & Llama-3 & 71.09&  \\
    % \hline
  \end{tabular}
  \caption{CIR Evaluation}
  \label{tab:manual-evaluation}
\end{table}

\section{Robustness of CIRE Across LLMs}
\label{sec:cire-robustness}
A limitation of our initial experiments was that CIRE was evaluated with only a single LLM (Llama-3.3). To strengthen this analysis, we repeated the evaluation using three additional models: Qwen-72B, Qwen-32B, and Llama-3.1-8B.
Results on the manually annotated subset (comparable to Table \ref{tab:lcire-evaluation}) are presented in Table \ref{tab:lcire-3models}. While there is some numerical variation, the relative rankings of de-identification systems remain consistent. \\
For the full dataset (comparable to Table \ref{tab:manual-evaluation}), results are shown in Table \ref{tab:manual-evaluation-3-models}.
Again, we observe stability in the relative performance of de-identification systems across LLMs.\\
Together, these results suggest that CIRE is robust to the choice of LLM used for evaluation, though further validation with additional models and datasets would be valuable.

\begin{table}[!htbp]
  \tiny
  \setlength{\tabcolsep}{2pt}
  \centering
  \begin{tabular}{llllll}
    \hline
    \textbf{Dataset} & \textbf{Evaluating Model} & \textbf{Model}  & \textbf{Accuracy} & \textbf{Precision} & \textbf{Recall}  \\
    \hline
    \multirow{9}{*}{AHS}
        & \multirow{3}{*}{Qwen 72B} & M. Deidentify & 0.94 &	0.95 &	0.42 \\
        & & CBERT + Presidio & 0.95 &	0.79 &	0.39 \\
        & & Llama-3.3 & 0.97 &	0.86	& 0.57\\
        \cline{2-6}
        & \multirow{3}{*}{Qwen 32B} & Deidentify  & 0.94 & 0.74 & 0.6 \\
        & & CBERT + Presidio & 0.96 &	0.86 &	0.43 \\
        & & Llama-3.3 & 0.98 &	0.92	& 0.57\\
        \cline{2-6}
        & \multirow{3}{*}{Llama 3.1 8B} & Deidentify  & 0.83 & 0.31 & 0.51 \\
        & & CBERT + Presidio & 0.84 &	0.28 &	0.79 \\
        & & Llama-3.3 & 0.88 &	0.06	& 0.1\\
    \hline
    \multirow{9}{*}{MIMIC-III}
        & \multirow{3}{*}{Qwen 72B} & M. Deidentify & 0.93 &	0.82 &	0.33 \\
        & & CBERT & 0.98 &	0.97 &	0.74 \\
        & & Llama-3.3 & 0.92 &	0.13 & 0.67\\
        \cline{2-6}
        & \multirow{3}{*}{Qwen 32B} & Deidentify  & 0.95 & 0.82 & 0.65 \\
        & & CBERT & 0.98 &	0.88 &	0.76 \\
        & & Llama-3.3 & 0.89 &	0.1	& 0.78\\
        \cline{2-6}
        & \multirow{3}{*}{Llama 3.1 8B} & Deidentify  & 0.74 & 0.22 & 0.62 \\
        & & CBERT & 0.78 &	0.23 &	0.89 \\
        & & Llama-3.3 & 0.81 &	0.05	& 0.56\\
     
    \hline
  \end{tabular}
  \caption{Results of CIRE with additional LLMs on the manually annotated subset (comparable to Table \ref{tab:lcire-evaluation}  M. Deidentify: Modified Deidentify; CBERT: ClinicalBERT}
  \label{tab:lcire-3models}
\end{table}

\begin{table}[!htbp]
\tiny
  \centering
  \begin{tabular}{llll}
    \hline
    \textbf{Dataset} & \textbf{Evaluating Model} &\textbf{Model} & \textbf{CIRE}  \\
    \hline
    \multirow{9}{*}{AHS}
        & \multirow{3}{*}{Qwen 72B} & Modified Deidentify & 0.92 \\
        & & ClinicalBERT + Presidio &  0.98\\
        & & Llama-3.3 & 0.94\\
        \cline{2-4}     
        & \multirow{3}{*}{Qwen 32B} & Modified Deidentify &0.89 \\
        & & ClinicalBERT + Presidio &  0.96\\
        & & Llama-3.3 & 0.94\\
        \cline{2-4}
        & \multirow{3}{*}{Llama 3.1 8B} & Modified Deidentify &0.87 \\
        & & ClinicalBERT + Presidio &  0.85\\
        & & Llama-3.3 & 0.95\\
  \hline
  \end{tabular}
  \caption{CIR Evaluation across different models}
  \label{tab:manual-evaluation-3-models}
\end{table}

\section{Chunking vs. Sentence Splitting}
\label{sec:chunking-vs-sentence}
The decision to employ fixed-length chunking rather than sentence-pair alignment was not a matter of design preference but of empirical necessity. Our private clinical dataset (AHS) was derived from non-native PDF sources using OCR, which introduced substantial formatting noise. In particular, sentence boundary detection was highly unreliable: in a manual review of 300 aligned sentences, only 159 were found to be valid sentences.
Given these limitations, sentence-level splitting would have further increased alignment errors and reduced the consistency of evaluation. For this reason, we employed fixed-length 20-token chunking, which offered a more reliable preprocessing strategy, though at the cost of occasionally dividing clinical concepts.\\
To evaluate the effect of this choice, we conducted a targeted experiment comparing sentence-level alignment to 20-token chunking on 100 samples each. Results showed near-identical performance between the two approaches: accuracy (0.980 vs. 0.983), precision (0.543 vs. 0.547), and recall (0.630 vs. 0.630).\\
These findings suggest that the alignment method (sentence-level vs. chunking) has minimal influence on model performance in this setting and that our decision to use chunking is justified by both practical robustness and empirical validation.

\section{Compute}
\label{sec:compute}
The experiments described in the paper were performed on two NVIDIA A100-SXM4-80GB GPUs. We roughly estimate that the total GPU time used was approximately 1550 hours.

Prompt-tuning took an approximate 100 GPU hours. De-identification took an approximate 1350 hours. Measuring the CIR metrics took an approximate 100 hours.

\section{Use of AI Assistants}
An AI assistant was used only for spelling, grammar, and phrasing.
\section{Instructions for Annotators}
\label{sec:annotators}
\begin{itemize}
  \item \textbf{Document Overview:}  
    This file lists the false positives produced by the de-identification model—that is, tokens removed unnecessarily that may carry important triage or diagnostic information. We sampled 500 of these tokens to categorize them and estimate the proportion of genuinely important or sensitive clinical data.

  \item \textbf{File Columns:}
    \begin{itemize}
      \item \textbf{File Name:} The source document in which the false positive occurred.
      \item \textbf{Edit Distance:} The Levenshtein distance between the original token and the de-identified token; lower values indicate minor changes (e.g., simple spelling fixes).
      \item \textbf{Original Token:} The token from the original text that was removed by the model.
      \item \textbf{De-identified Token:} The model’s replacement (or removal) of the original token.
      \item \textbf{Context:} A snippet of surrounding text, formatted as  
        \texttt{… / prev / prev / original\_token / deidentified\_token / next / next / …}.
      \item \textbf{Category:} Our assigned category for the original token (see below).
      \item \textbf{Severity of Change:} For clinically relevant removals, indicates whether the change is \emph{High} or \emph{Low} criticality.
    \end{itemize}

   \item \textbf{Categories:}
    \begin{itemize}
      \item \textbf{Clinically Relevant Changes:}  
        Removals or edits that alter key clinical information—medication names, dosages, critical terms—or that change meaning (e.g., “stop smoking”).
      \item \textbf{Clinically Irrelevant Changes:}  
        Tokens whose removal does not affect clinical interpretation (e.g., generic words like “it,” irrelevant dates, or “communication”).
      \item \textbf{Correct De-identification Missed by Human:}  
        Patient identifiers the manual review missed but the model correctly removed.
      \item \textbf{Insensitive Identifier:}  
        Non-sensitive identifiers (e.g., document IDs or encounter numbers) that the model removed.
      \item \textbf{Provider/Clinic Information:}  
        Details about providers or clinics—phone numbers, clinic names, addresses, provider IDs, etc.
      \item \textbf{Unknown:}  
        Tokens that do not clearly belong to any of the above categories and require your judgment.
    \end{itemize}

  \item \textbf{Reviewer Instructions:}
    \begin{itemize}
      \item \emph{Validate Categories:} Randomly inspect entries to confirm or correct the assigned \textbf{Category}.
      \item \emph{Resolve “Unknown” Tokens:} For every row labeled \textbf{Unknown}, choose the appropriate category.
      \item \emph{Set Severity for Relevant Changes:}  
        \begin{itemize}
          \item In the \textbf{Category} column, select \emph{Clinically Relevant Changes}.
          \item Enter \emph{High} or \emph{Low} in the \textbf{Severity of Change} column.
        \end{itemize}
    \end{itemize}
\end{itemize}

\section{Variation on De-identification}
\label{sec:deid-variations}
\subsection{Increasing PII to include provider information}
There is no consensus on the set of metrics that qualify as PII and need to be removed. For example, provider name is often not legally required to be removed. This was the approach we took in the main body of the text. However, as many models are trained to remove all names, our schema mis-match with that which they were trained on results in an under-estimation of their performance. To that end, we re-ran the evaluation on a subset of AHS (100 texts) considering provider information as PII. Results are presented in Table \ref{tab:clin-removed}.  We can see that the precision of all models improve, with substantial improvements for all models asides from Llama-3.3.

\subsection{Sub-token Modifications}
Likewise, there is no consensus on how much of a PII needs to be removed for the change to count as a true positive instead of a false negative. That is, if half a name is removed, should that count as being de-identified? In the work, we too the generous approach which considers any change in a token labelled as PII as being fully de-identified (i.e., a true positive). Table \ref{tab:standard-evaluation-conservative} presents the performance if we take the conservative approach instead. In the conservative approach, the whole set of characters belonging to the PII must be correctly classified to count as a true positive.

\newpage
\begin{table*}[!htbp]
\small
  \centering
  \begin{tabular}{lllllllllll}
    \hline
    \textbf{Dataset} & \textbf{Model} & \textbf{TP} & \textbf{TN} & \textbf{FP} & \textbf{FN} & \textbf{A} & \textbf{R} & \textbf{P} & \textbf{F1} &  \\
    \hline
    \multirow{4}{*}{AHS} & Presidio & 10025 &94154&16350&5492 & 0.83 &0.65 & 0.38 &  0.48

\\
    & Modified Presidio & 9463 &104689 &5815&6054 & 0.91 & 0.61& 0.62& 0.61 & \\
    & ClinicalBERT & 13708 & 96096& 14408&1809 &0.87& 0.88&0.49 & 0.63& \\
    & Modified ClinicalBERT &14023  &95847 &14657 &1494 & 0.87&0.90 & 0.49&0.63& \\
    & ClinicalBERT + Presidio & 14125 &94512 & 15992& 1392&0.86 &\textbf{0.91} &0.47 & 0.62& \\
    & Deidentify & 10217 &101511&8993 & 5300&0.89 &0.66 & 0.53&0.59& \\
    & Modified Deidentify & 11004 & 101117& 9387& 4513&0.89&0.71& 0.54 & 0.61\\
    & Llama-3.3 &12607  & 103213&7291&2910 & \textbf{0.92}& 0.81&\textbf{0.63} & \textbf{0.71} & \\
    \hline
  \end{tabular}
  \caption{AHS (clinic/provider identifiers removed) with standard classification metrics}
  \label{tab:clin-removed}
\end{table*}

\begin{table*}[ht]
\small
  \centering
  \begin{tabular}{lllllllllll}
    \hline
    \textbf{Dataset} & \textbf{Model} & \textbf{TP} & \textbf{TN} & \textbf{FP} & \textbf{FN} & \textbf{A} & \textbf{R} & \textbf{P} & \textbf{F1} &  \\
    \hline
    \multirow{4}{*}{AHS} & Presidio & 5436 & 120910 & 25879&1844 & 0.82 & 0.75& 0.17 &  0.28

\\
    & Modified Presidio & 5015 &133543 &13246 & 2265& 0.90& 0.69& 0.27&0.39 & \\
    & ClinicalBERT & 6468  & 121915&24874 &812 &0.83 &0.89 &0.21 &0.33 & \\
    & Modified ClinicalBERT & 6603 & 121386& 25403& 677& 0.83&0.91 & 0.21&0.34 & \\
    & ClinicalBERT + Presidio & 6643&119753 & 27036&637 &0.82 & 0.91& 0.20& 0.32 \\
    & Deidentify &5672 & 129649&17140 &1608 &0.88 &0.78 & 0.25& 0.38 \\
    & Modified Deidentify &6094 &128780 &18009 & 1186& 0.88&0.84 &0.25 & 0.39 \\
    & Llama-3.3 & 7214&144545 &2244 &66 &\textbf{0.99} &\textbf{0.99} &\textbf{0.76} & \textbf{0.86} \\
    \hline
    \multirow{4}{*}{MIMIC-III} &Presidio &109190 &2692556 &212497 &32778 &0.92 &0.77 & 0.34& 0.47 \\
    & Modified Presidio &104717  & 2883273&21780 &37251 &0.98 &0.74 &0.83 & 0.78& \\
    & ClinicalBERT & 124859 & 2890210&14843& 17109&\textbf{0.99}& 0.88& \textbf{0.89}&\textbf{0.89} & \\
    & Modified ClinicalBERT & 124951 &2887777& 17276&17017 &\textbf{0.99} &0.88 &0.88 &0.88& \\
    & ClinicalBERT + Presidio &110903 & 2863729& 41324&31065 &0.98 &0.78&0.73& 0.75 \\
    & Deidentify &114718 &2873971 & 31082&27250 & 0.98&0.81 &0.79 & 0.80 \\
    & Modified Deidentify & 114721& 2871704&33349 &27247 & 0.98& 0.81&0.77 & 0.79 \\
    & Llama-3.3 &133107 &2855168&49885&8861 &0.98 & \textbf{0.94}&0.73 &0.82  \\
    \hline
  \end{tabular}
  \caption{Conservative Evaluation}
  \label{tab:standard-evaluation-conservative}
\end{table*}

\begin{table*}[!htbp]
\tiny
  \centering
  \begin{tabular}{lllllll}
    \hline
    \textbf{Dataset} & \textbf{Model} & \shortstack{\textbf{Clinically Relevant}\\\textbf{Changes}} & \shortstack{\textbf{Clinically }\\\textbf{Irrelevant Changes}} & \shortstack{\textbf{provider/clinic}\\\textbf{information}} & \shortstack{\textbf{insensitive}\\\textbf{ identifier}}&\shortstack{\textbf{Correct de-id}\\\textbf{ missed by human }} \\
    \hline
    \multirow{3}{*}{AHS} 
      & Modified Deidentify      &    22	& 138	& 319	& 20&1    \\
      & ClinicalBERT + Presidio  &     14	& 134	& 328	& 22 & 2    \\
      & Llama-3.3 &     89	& 168	& 180	 &39&23     \\
    \hline
    \multirow{3}{*}{MIMIC-III} 
      & Deidentify              &     444 & 36 & 20 & 0 &0     \\
      & ClinicalBERT            &     263	& 67	 & 166	& 4&0     \\
      & Llama-3.3                 &    252 & 168 & 80 &0&0   \\
    \hline
  \end{tabular}
  \caption{Breakdown of De-identification Error Categories Across Models:
1. Clinically relevant changes: Modifications affecting medical meaning or patient care (e.g., drug-name corrections, dose removals or 'stop' in 'stop smoking').
2. Clinically irrelevant changes: Edits that do not alter clinical interpretation (e.g., removal of random dates or filler words like “thank,” “communication”).
3. Provider/clinic information: Redactions of healthcare facility or provider details (phone numbers, clinic names, addresses).
4. Insensitive identifiers: insensitive IDs (e.g., document or encounter numbers) that the model unnecessarily redacted.
5. Correct de-identifications missed by human: Patient identifiers the annotators missed but the model successfully redacted.}

  \label{tab:errors-classification}
\end{table*}

\end{document}